\documentclass[letterpaper, 10 pt, conference]{ieeeconf}  

\IEEEoverridecommandlockouts                              

\usepackage{graphics} 
\usepackage{amsmath} 
\usepackage{amssymb}  
\usepackage{graphicx}
\usepackage{pifont}
\usepackage[table]{xcolor}

\title{\LARGE \bf
Advancing Depth Anything Model for Unsupervised Monocular Depth Estimation in Endoscopy
}

\author{Bojian Li$^{1}$, Bo Liu$^{1}$, Xinning Yao$^{1}$, Jinghua Yue$^{1}$, and Fugen Zhou$^{1}$ 
\thanks{*This work was supported in part by the Beijing Natural Science Foundation (L222034, L232037, and L222104) and the National Natural Science Foundation of China (No. 12375359).}
\thanks{$^{1}$Bojian Li, Bo Liu, Xinning Yao, Jinghua Yue, and Fugen Zhou are with the Image Processing Center, Beihang University, Beijing, 100191, China. Corresponding Author: {\tt\small bo.liu@buaa.edu.cn}}%
}

\begin{document}

\maketitle
\thispagestyle{empty}
\pagestyle{empty}

\begin{abstract}
Depth estimation is a cornerstone of 3D reconstruction and plays a vital role in minimally invasive endoscopic surgeries. However, most current depth estimation networks rely on traditional convolutional neural networks, which are limited in their ability to capture global information. Foundation models offer a promising approach to enhance depth estimation, but those models currently available are primarily trained on natural images, leading to suboptimal performance when applied to endoscopic images. In this work, we introduce a novel fine-tuning strategy for the Depth Anything Model and integrate it with an intrinsic-based unsupervised monocular depth estimation framework. Our approach includes a low-rank adaptation technique based on random vectors, which improves the model's adaptability to different scales. Additionally, we propose a residual block built on depthwise separable convolution to compensate for the transformer's limited ability to capture local features. Our experimental results on the SCARED dataset and Hamlyn dataset show that our method achieves state-of-the-art performance while minimizing the number of trainable parameters. Applying this method in minimally invasive endoscopic surgery can enhance surgeons' spatial awareness, thereby improving the precision and safety of the procedures.

\end{abstract}

\section{INTRODUCTION}

Depth estimation is a crucial component of augmented reality navigation systems in minimally invasive endoscopic surgery, where accurate 3D depth information is essential for enhancing surgical precision and safety \cite{leonard2018evaluation,chu2018perception}. Given the difficulties in obtaining ground truth from endoscopic images and the fact that monocular endoscopes are more flexible and practical compared to stereo endoscopes, unsupervised monocular depth estimation (UMDE) algorithms have attracted broader research interest \cite{liu2024Self,zhou2017unsupervised}. These algorithms typically transform the depth estimation task into an image synthesis problem between adjacent viewpoints, using image synthesis error to guide network training \cite{godard2019digging}. However, conventional depth estimation algorithms face significant challenges when applied to endoscopic images due to factors such as lighting variations and sparse textures \cite{li2024image}.

To address the challenges of UMDE in endoscopy, particularly those related to lighting variations, several approaches have been proposed. For instance, AF-SfMLearner \cite{shao2022self} designs an appearance flow extraction network for nonlinear adjustment of image lighting. IID-SfMLearner \cite{li2024image} combines image intrinsic decomposition with depth estimation, incorporating reflectance invariance to supervise network training. These strategies have significantly improved the performance of monocular endoscopic depth estimation. However, the depth estimation networks in these methods rely on simple convolutional neural networks, which primarily capture local information and are limited in their ability to perceive global information.

The emergence of foundation models has sparked a new revolution in the field of computer vision \cite{kolides2023artificial,kirillov2023segment}. These models, often based on Vision Transformers (ViT), excel at effectively extracting global image features. For dense prediction tasks like semantic segmentation and depth estimation, having a global receptive field for each pixel relative to the scene allows the network to estimate local information with greater accuracy \cite{agarwal2022depthformer}. Moreover, these models are typically trained on extremely large-scale datasets, endowing them with robust zero- and few-shot capabilities across a wide range of downstream scenarios \cite{radford2021learning}.

\begin{figure}[!t]
\centerline{\includegraphics[width=\columnwidth]{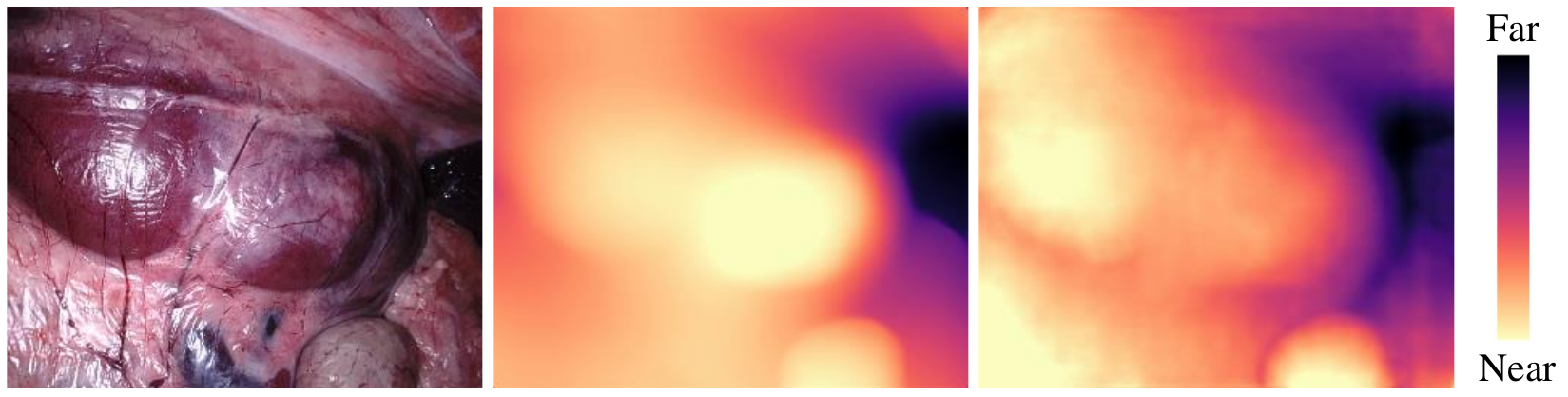}}
\caption{Left: endoscopic image from minimally invasive surgery scenes. Middle: depth estimation result from Depth Anything Model \cite{yang2024depth}. Right: more accurate depth estimation result from our model.}
\label{fig1_1}
\end{figure}

\begin{figure*}[!t]
\centerline{\includegraphics[width=1\linewidth]{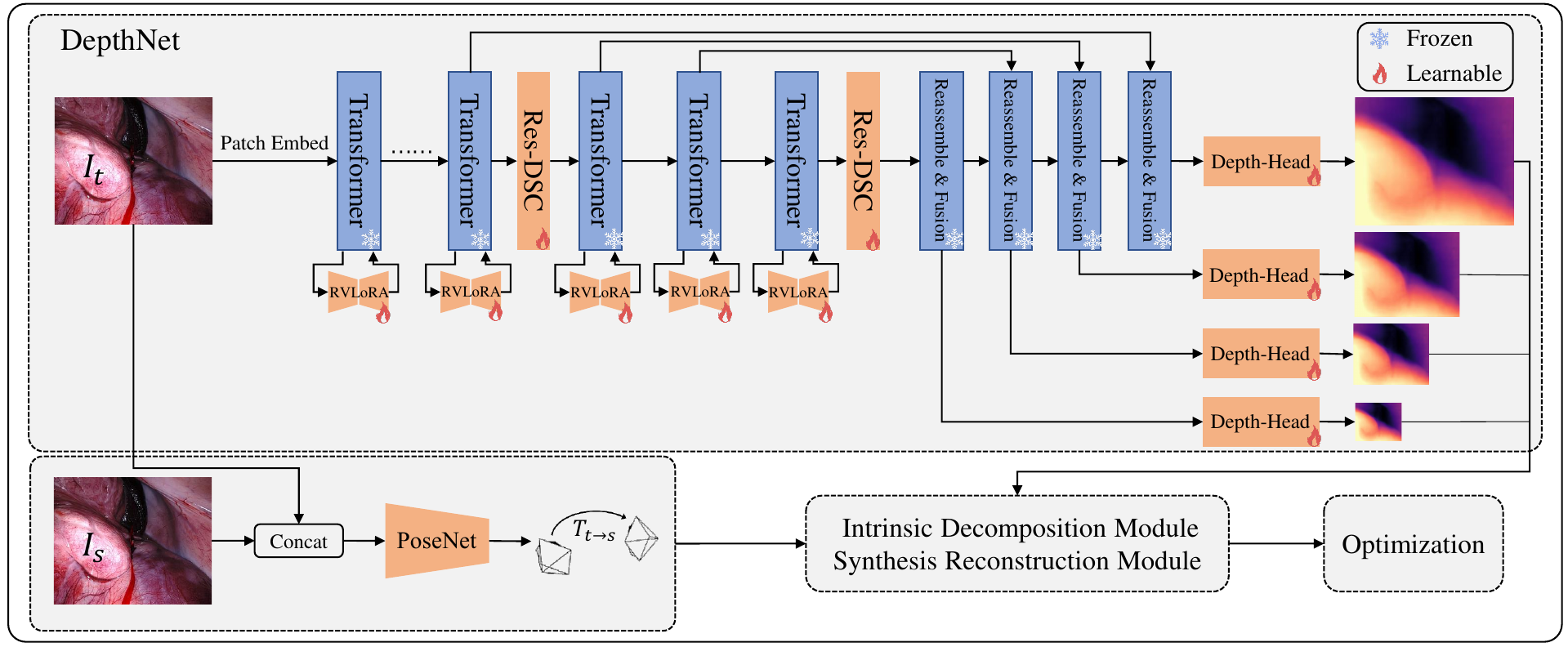}}
\caption{Network architecture diagram. The DepthNet is used to predict the depth information. The PoseNet is used to estimate the camera’s pose. The intrinsic decomposition module is used to decompose the images into reflectance and shading components. The synthesis reconstruction module is used to reconstruct the target image. The training method is the same with the IID-SfMLearner \cite{li2024image}. The encoder in the depth network contains 12 transformer modules, each with an RVLoRA module. Res-DSC modules are added after the 3rd, 6th, 9th, and 12th layers.}
\label{fig1}
\end{figure*}

Recently, a foundation model (Depth Anything Model) for depth estimation in natural images has been proposed \cite{yang2024depth}. However, as shown in Fig. \ref{fig1_1}, this model exhibits significant performance degradation when directly applied to endoscopic images \cite{cui2024endodac}. Furthermore, due to the scarcity of annotated endoscopic data, training a foundation model for endoscopic depth estimation from scratch is impractical. Therefore, fine-tuning the Depth Anything Model for specific endoscopic scenarios has become an effective approach to addressing this issue.

There are numerous strategies for fine-tuning foundation models, with Low-Rank Adaptation (LoRA) being one of the most commonly used methods \cite{hu2021lora}. However, LoRA does not always perform optimally and some variations show better performance in certain tasks \cite{fang2024dropout}. To address this, we explore a novel fine-tuning strategy aimed at enhancing performance without increasing the number of parameters in LoRA. Inspired by recent studies that highlight the surprising effectiveness of models using random weights and projections \cite{kopiczko2023vera, lu2022frozen}, we propose a low-rank adaptation method based on random vectors (RVLoRA) that enhances model performance without adding additional trainable parameters. Additionally, transformer models tend to focus more on low-frequency information, resulting in insufficient ability to extract texture details and local features \cite{wu2021cvt,si2022inception}. To address this, we propose a residual block based on depthwise separable convolution (Res-DSC) to achieve higher performance by enhancing locality.

In summary, our main contributions are as follows:
\begin{itemize}
\item We designed a novel fine-tuning strategy, RVLoRA, for depth estimation foundation model. This strategy enhances the model's adaptability to different scales by introducing random scaling vectors, without adding additional trainable parameters.

\item We introduced the Res-DSC module, which integrates a CNN component into the Vision Transformer (ViT), thereby improving the model’s ability to capture local features.

\item Experiments conducted on SCARED dataset and Hamlyn dataset demonstrate that our approach achieves state-of-the-art results with the minimal trainable parameters.
\end{itemize}

\section{RELATED WORKS}
\subsection{Unsupervised Monocular Depth Estimation}
UMDE has been effectively explored on natural images. Zhou et al. \cite{zhou2017unsupervised} were the first to propose using view synthesis for UMDE. Subsequently, Godard et al. \cite{godard2019digging} further enhanced the performance by utilizing per-pixel loss minimization and automatic masking. Zheng et al. \cite{Zheng2023STEPS} improved depth estimation in nighttime scenarios by jointly learning image enhancement and depth estimation.

Despite the advances these methods have made for natural images,  most of them rely on the assumption of photometric consistency, which renders them less effective in endoscopic scenarios with varying illumination. To address this, Ozyoruk et al. \cite{ozyoruk2021endoslam} and Shao et al. \cite{shao2022self} first adjusted the lighting of images before performing depth estimation. Li et al. \cite{li2024image} improved the results by employing intrinsic decomposition to separate illumination variations and using reflectance consistency. Besides this line of research, Batlle et al. \cite{Batlle2022Photometric} used photometric stereo technique to achieve depth estimation of the human colon, but this method only performed well on the cavitary endoscope dataset. Cui et al. \cite{cui2024endodac} combined foundation model techniques for depth estimation, achieving relatively better results.

\subsection{Low-Rank Adaptation}
LoRA \cite{hu2021lora} marked a significant breakthrough in efficiently training large language models for specific tasks by decomposing matrices into two low-rank matrices, A and B, thereby drastically reducing training resource consumption. Based on this idea, several improvements have been proposed recently. For example, LoRA+ \cite{hayou2024lora} further enhanced efficiency by introducing different learning rates for matrices A and B. Zhang et al. \cite{zhang2023adaptive} introduced AdaLoRA, which selects different ranks for different adapters. This method effectively enhances LoRA's learning capability and training stability. VeRA \cite{kopiczko2023vera} reduced the parameter count even further by freezing matrices A and B and introducing additional vectors for training, though this led to a slight loss in accuracy. Different fine-tuning strategies exhibit varying performance across different tasks. Therefore, we need to explore a fine-tuning strategy that is suitable for depth estimation tasks.

\section{METHOD}
In this section, we first review the framework of image intrinsic-based unsupervised monocular depth estimation, which serves as the baseline for our method. We then describe how we fine-tune the foundational model to obtain a better depth estimation. Fig. \ref{fig1} illustrates our network framework.

\subsection{Image Intrinsic-Based UMDE}
In classic UMDE framework \cite{zhou2017unsupervised}, taking one frame as the target image $I_t$ and adjacent frames as source images $I_s$, a depth estimation network is used to estimate the depth information $D_t$ of the target image, and a pose estimation network is used to estimate the camera's pose transformation matrix $T_{t\rightarrow s}$ between adjacent frames. Using known camera intrinsic matrix $K$, the position correspondence between the $I_t$ and $I_s$ is determined:
$$
 p_s \sim K T_{t \rightarrow s} D_t\left(p_t\right) K^{-1} p_t \eqno{(1)}
$$
here, $p_t$ and $p_s$ represent the pixel coordinates of the same point in three-dimensional space under the target view and the source view, respectively. Based on these, a synthesized frame $I_{s\rightarrow t}$ can be reconstructed via image warping. And the photometric loss between the reconstructed image $I_{s\rightarrow t}$ and the real target image $I_t$ is calculated to supervise network training which is commonly constructed as a weighted combination of L1 loss and SSIM loss.

This framework relies on the assumption of photometric consistency, which no longer holds in endoscopy. IID-SfMLearner \cite{li2024image} employs intrinsic image decomposition to tackle the challenge, making it the most advanced framework for monocular endoscopic depth estimation. IID-SfMLearner additionally introduces the intrinsic decomposition module and synthetic reconstruction module to decompose the image into reflectance and shading images. This allows the method to leverage the inherent reflectance consistency of images, providing better supervision during network training.

\subsection{Advancing Depth Anything Model for UMDE}
The depth estimation networks typically employed in IID-SfMLearner and other classic UMDE methods are standard convolutional neural networks, which lack sufficient global perception capabilities for accurately capturing depth information. Fine-tuning a foundation model is a straightforward approach to enhance long-range dependencies without significantly increasing the number of trainable parameters. Although the Depth Anything Model \cite{yang2024depth} has been proposed for depth estimation, it was trained on natural images, and applying it directly to endoscopic images leads to poor performance. To address this, we propose a novel fine-tuning strategy specifically tailored for endoscopic images. Our approach introduces two key components: low-rank adaptation based on random vectors and residual blocks utilizing depthwise separable convolution.

\subsubsection{Low-rank Adaptation Based on Random Vectors (RVLoRA)}
As models grow larger, full fine-tuning becomes increasingly impractical. Neural networks consist of many dense layers that perform matrix multiplications, and the weight matrices in these layers are typically full-rank. However, when adapting to specific tasks, pre-trained foundation models exhibit low "intrinsic dimensions" \cite{aghajanyan2021intrinsic}. This means they can still learn effectively even when their parameters are projected into smaller, random subspaces. Inspired by this, LoRA was proposed, assuming that weight updates during adaptation also have low "intrinsic rank" \cite{hu2021lora}. For a pre-trained weight matrix $W_0\in\mathbb{R}^{m\times n}$, its update can be represented using low-rank decomposition as $W_0+\Delta W=W_0+BA$, and the modified forward pass is:
$$
h=W_{0}x+\Delta Wx=W_{0}x+BAx \eqno{(2)}
$$
where $B\in\mathbb{R}^{m\times r}$ and $A\in\mathbb{R}^{r\times n}$, with rank $r \ll \min (m, n)$. During training, $W_0$ is frozen and does not receive gradient updates, while $A$ and $B$ are trainable. In this way, the pre-trained model weights are frozen, and the trainable low-rank decomposition matrices are injected into each layer of the transformer architecture, significantly reducing the number of trainable parameters for downstream tasks.

\begin{figure}[!t]
\centerline{\includegraphics[width=\columnwidth]{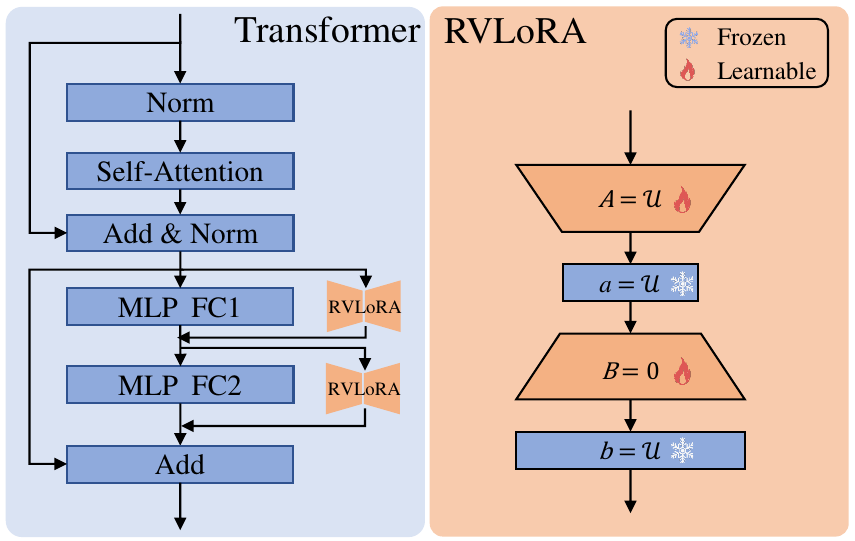}}
\caption{The deployment and detailed structure of the RVLoRA. RVLoRA is connected to the two linear layers of the MLP in each transformer module. The parameters a and b are frozen while A and B are trainable.}
\label{fig2}
\end{figure}

\begin{figure}[!t]
\centerline{\includegraphics[width=\columnwidth]{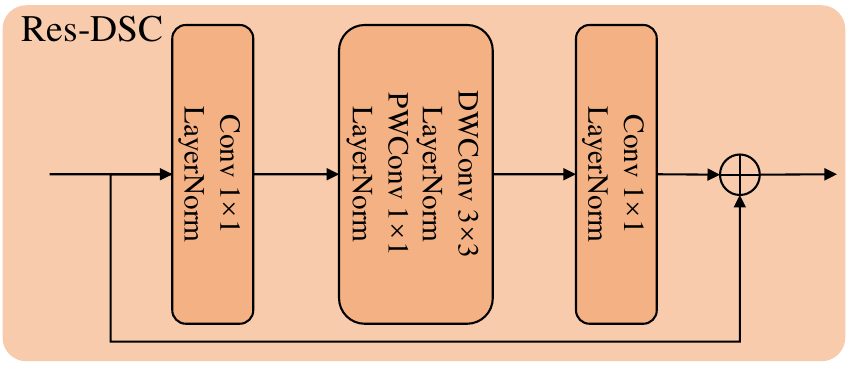}}
\caption{The detailed structure of the Res-DSC module. Res-DSC contains a depthwise separable convolution, a pointwise separable convolution, and the residual connection.}
\label{fig3}
\end{figure}

Recent studies have demonstrated the remarkable effectiveness of using random weights in deep learning models. On the one hand, introducing random vectors can significantly enhance a model's adaptive capability, providing an effective mechanism for adjusting the model's behavior without the need for extensive retraining or parameter tuning \cite{kopiczko2023vera, lu2022frozen}. On the other hand, several advanced deep learning techniques, including those focused on feature learning and transformations, rely heavily on affine transformations of learned features. Freezing these randomly initialized parameters has been shown to improve the network's ability to capture and generalize these affine transformations more effectively \cite{frankle2021training}. Additionally, in monocular depth estimation tasks, an inherent challenge is scale ambiguity, as depth information can vary in different contexts. Introducing random vectors specifically for scale adjustment can greatly improve the model's flexibility, enabling it to better adapt to varying scales during inference. Inspired by these insights, we propose a novel and parameter-efficient fine-tuning method: low-rank adaptation based on random vectors (RVLoRA). 

Fig. \ref{fig2} illustrates the detailed structure of RVLoRA. This method builds on LoRA by adding randomly initialized vectors $a$ and $b$, scaling the low-rank matrix to enhance the model's adaptability to different scales. During training, we employ Kaiming uniform initialization \cite{he2015delving} for  $a$, $b$, and $A$ while the low-rank matrices $B$ are initialized to zeros to ensure that the weight matrix remains unaffected during the first forward pass. The frozen scaling vectors and the trainable low-rank matrices are combined with the original weights without incurring additional inference delay. Our method can be represented as:
$$
h=W_0 x+\Delta W x=W_0 x+\Lambda_b B \Lambda_a A x\eqno{(3)}
$$
where $B\in\mathbb{R}^{m\times r}$, $A\in\mathbb{R}^{r\times n}$, $\Lambda_b\in\mathbb{R}^{m\times m}$, $\Lambda_a\in\mathbb{R}^{r\times r}$, with rank $r \ll \min (m, n)$. In this method, the low-rank matrices $B$ and $A$ are trainable, while the scaling vectors $b\in\mathbb{R}^{m\times 1}$ and $a\in\mathbb{R}^{r\times 1}$ are randomly initialized and remain frozen during training. These vectors are formally represented as diagonal matrices $\Lambda_b$ and $\Lambda_a$. This approach effectively performs random scaling on the rows of the low-rank matrices $B$ and $A$, thereby enhancing the model’s adaptability across layers.

\subsubsection{Residual Block Based on Depthwise Separable Convolution (Res-DSC)}

Transformers offer advantages like dynamic attention, global context awareness, and strong generalization capabilities. However, they often struggle to extract fine details and local features. On the other hand, convolutions excel at capturing local features and come with inherent benefits such as shift, scale, and distortion invariance, though they are limited in their ability to perceive global context. Therefore, integrating CNNs with Transformers can create a synergistic effect, significantly enhancing the overall model performance.

Inspired by this, we designed a residual block based on depthwise separable convolution (Res-DSC), as illustrated in Fig. \ref{fig3}. This structure consists of three convolutional layers: the first layer is a 1$\times$1 convolution to reduce the number of channels; the second layer is a 3$\times$3 depthwise separable convolution which has fewer parameters, leading to not only faster computation but also a more streamlined model compared to traditional convolution; the final layer is a 1$\times$1 convolution to restore the number of channels. To reduce the model's parameters while ensuring  the accuracy of depth estimation, we only add Res-DSC after the 3rd, 6th, 9th, and 12th transformer blocks. In Fig. \ref{fig1}, only the Res-DSC blocks after the 9th and 12th transformer blocks are shown, while the others are omitted for brevity.

\subsubsection{Decoder}
The decoder part adopts the same structure as Depth Anything Model \cite{yang2024depth}. The features from the last four layers of the encoder are fed into the decoder. Through the frozen Reassemble $\&$ Fusion module in the decoder, features at different resolutions are obtained and then aggregated. These aggregated features are then fed into the trainable Depth-Head module for the final dense prediction, resulting in depth estimation outputs at four scales.

\subsection{Total Training Framework}
We integrated our depth estimation network within the IID-SfMLearner framework \cite{li2024image}. In this framework, the intrinsic decomposition module performs intrinsic decomposition of the images, using decomposing-synthesis loss $L_{ds}$ and albedo loss $L_a$ to ensure the accuracy of the decomposition results and reflectance consistency between adjacent frames. The synthesis reconstruction module flexibly adjusts the illumination map and utilizes the decomposition results of the source image to reconstruct the target image, introducing mapping-synthesis loss $L_{ms}$ to supervise network training.

Besides, we apply an edge-aware depth smooth loss $L_{es}$ \cite{godard2017unsupervised} to enhance smoothness in non-edge regions while preserving sharp edges and details:
$$
L_{es}(D_t,I_t)=|\partial_x D_t|e^{-|\partial_x I_t|}+\left|\partial_y D_t\right|e^{-|\partial_y I_t|} \eqno{(4)}
$$
where $\partial D_t$ and $\partial I_t$ are the first derivative of the depth map $D_t$ and the target image $I_t$. 
The final loss function is defined as:
$$
loss =\lambda_{ds}L_{ds} + \lambda_aL_a +\lambda_{ms}L_{ms}+\lambda_{es} L_{es} \eqno{(5)}
$$
where $\lambda_{ds}$, $\lambda_a$, $\lambda_{ms}$ and $\lambda_{es}$ are weighting factors.

\begin{table}[t]
  \centering
    \caption{The calculation of applied evaluation metrics, in which $d$ represents the prediction depth and $d^*$ represents the ground truth.}
    \label{tab1}
    \resizebox{0.45\textwidth}{!}{
    \begin{tabular}{c | c} 
    \hline
    \textbf{Metric}	&\textbf{Definition}\\
    \hline 
       Abs Rel & $\frac{1}{|N|}\sum_{d\in N}|d-d^*|/d^*$\\
      Sq Rel & $\frac{1}{|N|}\sum_{d\in N}|d-d^*|^2/d^*$\\
      RMSE & $\sqrt{\frac{1}{|N|}\sum_{d\in N}(d-d^*)^2}$\\
      RMSE log & $\sqrt{\frac{1}{|N|}\sum_{d\in N}(\ln d-\ln d^{*})^{2}}$\\
      $\delta$  & $\frac{1}{|N|}\left\{d \in N \left\lvert\, \max \left(\frac{d}{d^*}, \frac{d^*}{d}\right)<1.25\right.\right\} \times 100 \%$\\
    \hline
    \end{tabular}
    }
\end{table}

\begin{figure*}[!t]
\centerline{\includegraphics[width=0.95\linewidth]{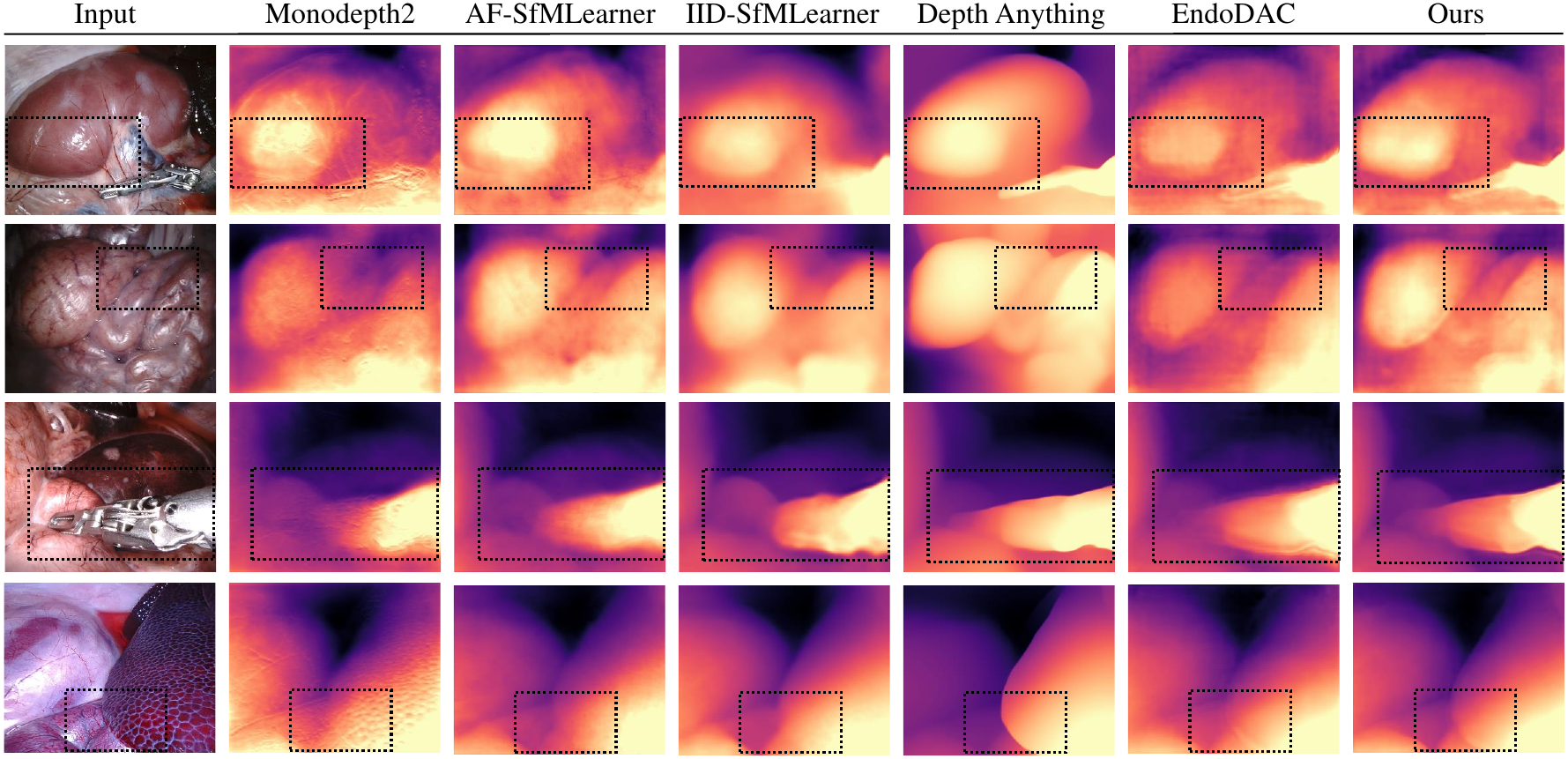}}
\caption{The visual results of the comparative experiments. It is evident that our method performs better in capturing details such as object edges. In particular, the regions highlighted by the black boxes demonstrate a significant advantage of our approach.Note:Since the ground truth in this dataset is captured using structured light encoding, comparing it with the depth map visually is not intuitive. As with other papers \cite{shao2022self,li2024image,cui2024endodac}, we do not present the ground truth for comparison.}
\label{fig4}
\end{figure*}

\begin{table*}[t]
  \centering
    \caption{Comparative experimental results on the SCARED dataset. The best results are presented in bold.}
    \label{tab2}
    \resizebox{1.0\textwidth}{!}{
    \begin{tabular}{c|c c c c c |c c } 
    \hline
      Method&Abs Rel$\downarrow$&Sq Rel$\downarrow$&RMSE$\downarrow$&RMSE log$\downarrow$&$\delta \uparrow$&Total.(M)&Train.(M)\\
        \hline 
        SfMLearner \cite{zhou2017unsupervised} & 0.086$\pm$0.037 & 1.021$\pm$1.100 & 7.553$\pm$5.110 & 0.121$\pm$0.060 & 0.925$\pm$0.081 & 31.60 & 31.60 \\
       Monodepth2 \cite{godard2019digging} & 0.066$\pm$0.025 & 0.577$\pm$0.482 & 5.781$\pm$3.174 & 0.093$\pm$0.036 & 0.961$\pm$0.043 & 14.84 & 14.84\\
       Endo-SfMLearner \cite{ozyoruk2021endoslam} & 0.068$\pm$0.029 & 0.679$\pm$0.709 & 6.227$\pm$4.137 & 0.098$\pm$0.047 & 0.955$\pm$0.054 & 14.84 & 14.84\\
       AF-SfMLearner \cite{shao2022self} & 0.060$\pm$0.024 & 0.443$\pm$0.347 & 4.964$\pm$2.348 & 0.082$\pm$0.032 & 0.973$\pm$0.037 & 14.84 & 14.84\\
        IID-SfMLearner \cite{li2024image} & 0.058$\pm$0.026  & 0.435$\pm$0.375  & 4.820$\pm$2.583  & 0.080$\pm$0.035  & 0.969$\pm$0.043 & 14.84 & 14.84\\ 
         Depth Anything \cite{yang2024depth} & 0.086$\pm$0.039  & 0.927$\pm$0.875  & 6.957$\pm$3.817  & 0.112$\pm$0.049  & 0.929$\pm$0.089 & 97.50 & 97.50\\ 
         EndoDAC \cite{cui2024endodac} & 0.051$\pm$0.022  & 0.355$\pm$0.334  & 4.442$\pm$2.509  & 0.073$\pm$0.032  & 0.979$\pm$0.032 & 99.09 & 1.66\\ 
        Ours &\textbf{0.048$\pm$0.021} &\textbf{0.315$\pm$0.297} & \textbf{4.172$\pm$2.261} & \textbf{0.068$\pm$0.030} & \textbf{0.982$\pm$0.027} & 98.80 & \textbf{1.38}\\
        \hline
     \end{tabular}
    } 
   
\end{table*}

\section{EXPERIMENT}
\subsection{Implementation Details}
All models are implemented using PyTorch \cite{paszke2019pytorch} and trained end-to-end using the Adam optimizer \cite{kingma2014adam} with $\beta_1 = 0.9$ and $\beta_2 = 0.999$. We train for 30 epochs on a GeForce RTX 3090 GPU with a batch size of 8. The initial learning rate is set to $1\times 10^{-4}$ and is multiplied by a scale factor of 0.1 after 10 epochs. In our experiments, the rank for RVLoRA is set to 4, the loss weights $\lambda_{ds}$, $\lambda_a$, $\lambda_{ms}$, and $\lambda_{es}$ are set to 0.2, 0.2, 1, and 0.01, respectively.
\subsection{Dataset and Metrics}
We conducted our experiments on the SCARED dataset, which was collected using the da Vinci Xi endoscope during the dissection of fresh pig abdomens \cite{allan2021stereo}. High-quality depth ground truth was obtained using structured light encoding during the acquisition process. Following the splitting strategy in prior works  \cite{li2024image,shao2022self}, we divided the SCARED dataset into 15,351 frames for training, 1,705 frames for validation, and 551 frames for testing. 

We adopted commonly used evaluation metrics such as Abs Rel, Sq Rel, RMSE, RMSE log, and Threshold $\delta$, with their calculation formulas shown in Table \ref{tab1}. Similar to previous methods \cite{li2024image,shao2022self}, during the evaluation, we scale the predicted depth map via the median scaling, which can be expressed as:
$$
  D_{scaled}=d*\left(median(d^{*})/median(d)\right) \eqno{(6)}
$$
where $d$ and $d^{*}$ represents the predicted depth and the ground truth, respectively. The upper limit for scaling the depth map is set to 150 millimeters, which effectively covers almost all pixels in the SCARED dataset.

\begin{figure*}[!t]
\centerline{\includegraphics[width=0.95\linewidth]{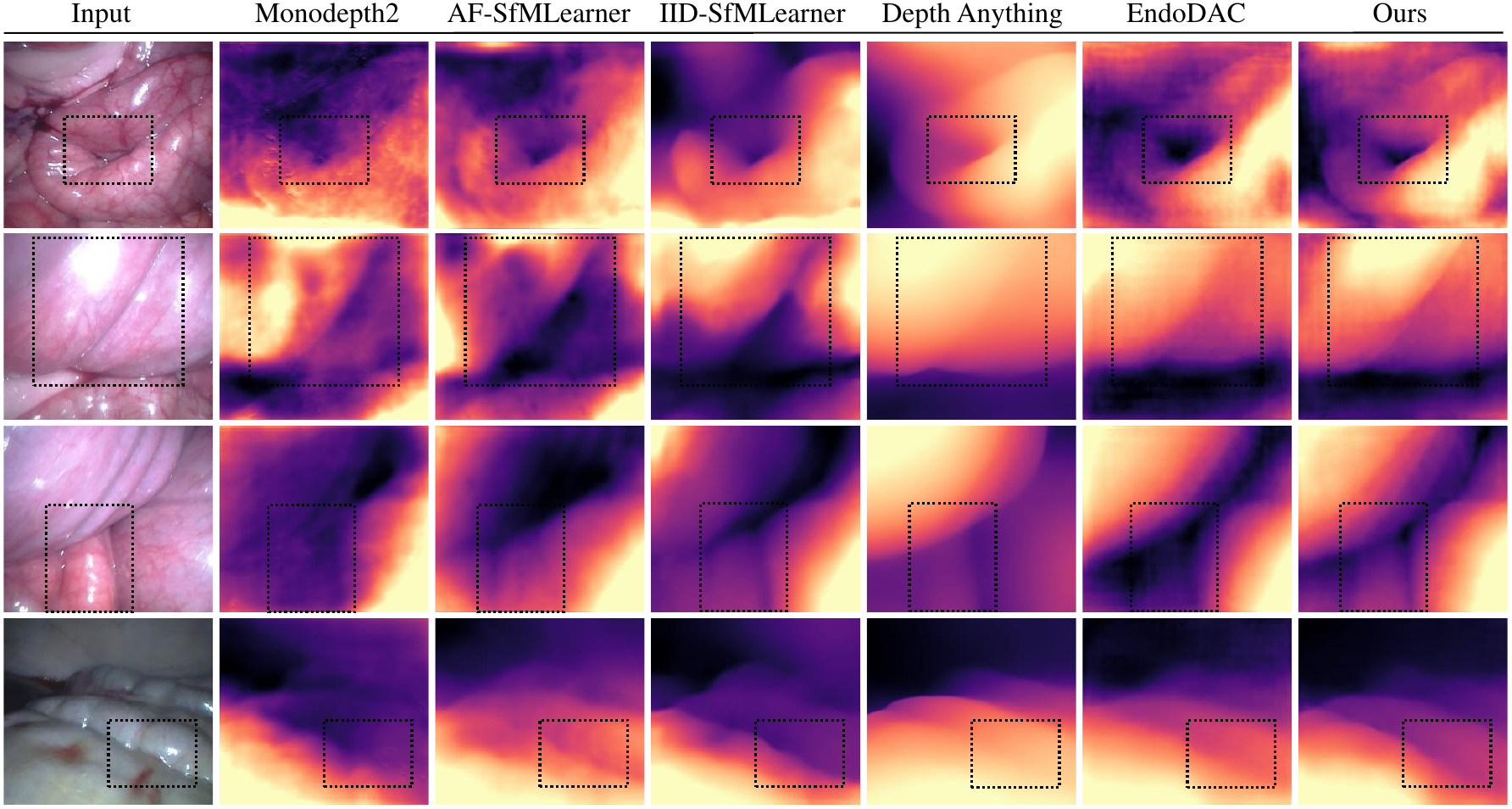}}
\caption{Qualitative comparison of applying the model trained on the SCARED dataset to the Hamlyn dataset for validation without any fine-tuning. Our method not only avoids erroneous estimates but also delivers more precise details. The regions highlighted by the black boxes illustrate a notable advantage of our approach, further demonstrating its strong generalization ability.}
\label{fig6}
\end{figure*}

\subsection{Comparison Experiments}
We compared our method with seven competitive unsupervised depth estimation methods: SfmLearner \cite{zhou2017unsupervised}, Monodepth2 \cite{godard2019digging}, Endo-SfmLearner \cite{ozyoruk2021endoslam}, AF-SfmLearner \cite{shao2022self}, IID-SfmLearner \cite{li2024image}, Depth Anything \cite{yang2024depth}, and EndoDAC \cite{cui2024endodac}. Among these, EndoDAC is currently the state-of-the-art method on the SCARED dataset. The first three methods are reproduced from scratch following the code provided by the authors, and the last four methods are tested using the optimal model provided by the authors. Table \ref{tab2} presents the experimental results and parameter numbers of the comparative experiments.

SfMLearner and Monodepth2, which are designed for natural scenes, perform poorly on endoscopic images. While Endo-SfMLearner, AF-SfMLearner, and IID-SfMLearner have introduced algorithmic improvements tailored to the unique characteristics of endoscopic images, their performance gains remain limited. Depth Anything Model, the latest foundation model for depth estimation, also falls short due to the absence of medical images in its training datasets. EndoDAC, which similarly fine-tunes the Depth Anything Model, has achieved significant improvements over earlier methods. Compared to EndoDAC, our method improves the fine-tuning approach and reduces the number of trainable parameters. In comparison, our method outperforms all others across various metrics. Notably, our approach has the fewest trainable parameters, totaling only 1.38M, which accounts for just 1.4\% of the overall parameters.

Fig. \ref{fig4} presents some representative results for the compared methods. It can be observed that the depth maps estimated by our method are smoother and more continuous, with better performance at object boundaries, indicating that our method exhibits superior performance in capturing image detail information. 

Additionally, we compared our method with other fine-tuning strategies, and the results are presented in Table \ref{tab3}. It is evident that our approach provides the optimal fine-tuning for the depth estimation foundation model.

\begin{table}[t]
  \centering
    \caption{The comparison of fine-tuning strategies}
    \label{tab3}
    \resizebox{0.48\textwidth}{!}{
    \begin{tabular}{c |c c c c c } 
    \hline
    Fine-tuning strategy & Abs Rel$\downarrow$ & Sq Rel$\downarrow$ & RMSE$\downarrow$ & RMSE log$\downarrow$ & $\delta \uparrow$ \\ \hline
        LoRA \cite{hu2021lora} & 0.051 & 0.349 & 4.383 & 0.072 & 0.978  \\
        VeRA \cite{kopiczko2023vera}  & 0.054 & 0.404 & 4.706 & 0.076 & 0.977  \\ 
        DVLoRA \cite{cui2024endodac} & 0.050 & 0.340 & 4.323 & 0.071 & 0.981 \\ 
        Ours &\textbf{0.048} &\textbf{0.315} & \textbf{4.172} & \textbf{0.068} & \textbf{0.982}  \\
        \hline
     \end{tabular}
    } 
   
\end{table}

\begin{table}[t]
  \centering
    \caption{Experimental results on the Hamlyn dataset. The best results are presented in bold.}
    \label{tab4}
    \resizebox{0.48\textwidth}{!}{
    \begin{tabular}{c|c c c c c } 
    \hline
      Method& Abs Rel$\downarrow$&Sq Rel$\downarrow$&RMSE$\downarrow$&RMSE log$\downarrow$&$\delta \uparrow$\\
        \hline 
       AF-SfMLearner \cite{shao2022self} & 0.130 & 2.134 & 11.306 & 0.165 & 0.843 \\
    IID-SfMLearner \cite{li2024image} & 0.120  & 1.833  & 10.362  & 0.156  & 0.867 \\ 
        EndoDAC \cite{cui2024endodac}  & 0.103  & 1.422  & 9.082  & 0.130  & 0.906 \\ 
        Ours & \textbf{0.094} &\textbf{1.215} & \textbf{8.318} & \textbf{0.120} & \textbf{0.924} \\
        \hline
     \end{tabular}
    } 
   
\end{table}

Like other methods \cite{shao2022self}, to further test the generalization ability of the proposed method, we directly applied the model trained on SCARED to the Hamlyn dataset for validation without any fine-tuning. The Hamlyn dataset is a laparoscopic/endoscopic video dataset created by the Hamlyn Centre at Imperial College London, containing rich laparoscopic and endoscopic video data \cite{David2021endo}. 

The test results on Hamlyn dataset are shown in the Table \ref{tab4}. We only compare with a few of the most competitive methods. It can be seen that our method demonstrates strong generalization ability and continues to achieve state-of-the-art performance on this dataset. The visual results, as shown in the Fig. \ref{fig6}, clearly illustrate that the depth maps generated by our method exhibit greater continuity and superior performance in capturing object boundaries and texture details. This experiment provides strong evidence of the robust generalization capability of our approach.

\begin{figure*}[!t]
\centerline{\includegraphics[width=0.8\linewidth]{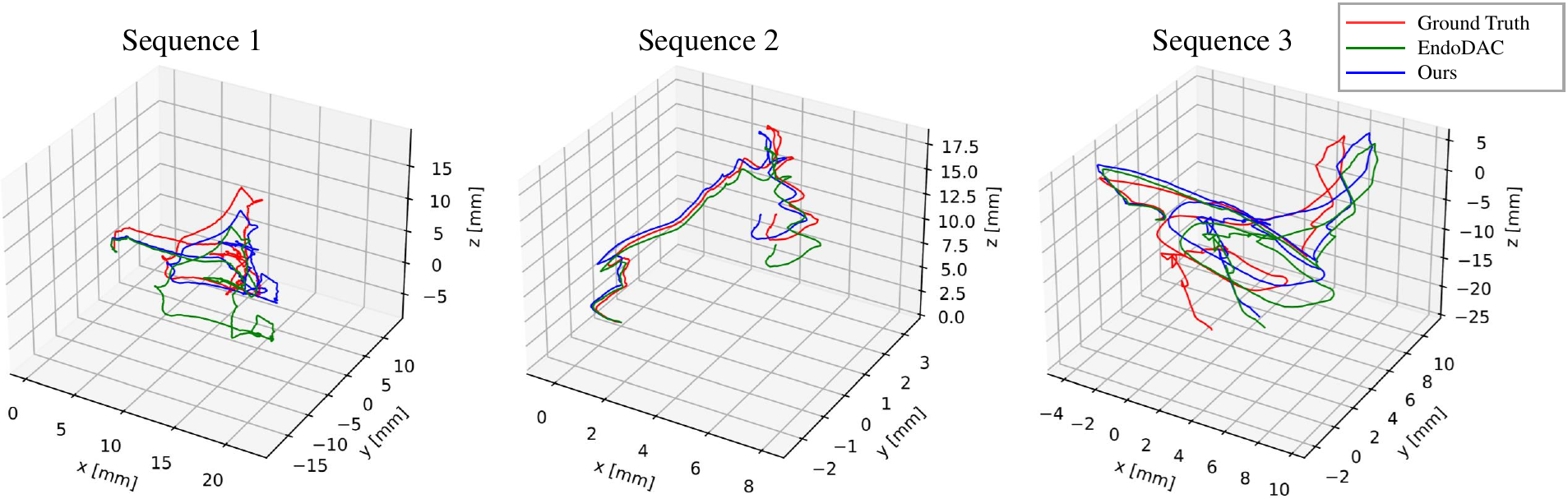}}
\caption{Visualization of pose estimation results. The red line represents the ground truth, the green line represents the result from EndoDAC, and the blue line represents our method's result. It can be observed that our results exhibit less trajectory drift and align more closely with the ground truth.}
\label{fig5}
\end{figure*}

\subsection{Ablation Study}
To further demonstrate the effectiveness of the proposed modules, ablation experiments were conducted and the results are shown in Table \ref{tab5}. The first row in the table represents the results obtained by directly testing the Depth Anything Model without any fine-tuning. The subsequent rows show the results of progressively adding the proposed two modules. The experimental results indicate that both modules effectively improve the depth estimation performance.

Compared to other initialization methods of the frozen parameters, Table \ref{tab8} shows the effectiveness of Kaiming uniform initialization.

In our method, the Res-DSC modules are evenly distributed across the 12 transformer modules of the encoder. For instance, when using three Res-DSC modules, they are placed at the 4th, 8th, and 12th layers. To determine the optimal number of Res-DSC modules, we conducted a series of ablation experiments. The results, shown in Table \ref{tab6}, indicate that using four Res-DSC modules achieves the best performance while maintaining a relatively low parameter count.

\begin{table}[t]
  \centering
    \caption{The ablation study results.}
    \label{tab5}
    \resizebox{0.48\textwidth}{!}{
    \begin{tabular}{c c |c c c c c } 
    \hline
    RVLoRA & Res-DSC & Abs Rel$\downarrow$ & Sq Rel$\downarrow$ & RMSE$\downarrow$ & RMSE log$\downarrow$ & $\delta \uparrow$ \\ \hline
        \ding{53} & \ding{53} & 0.086 & 0.927 & 6.957 & 0.112 & 0.929  \\ 
        \checkmark & \ding{53} & 0.052 & 0.351 & 4.382 & 0.074 & 0.977  \\ 
        \ding{53} & \checkmark & 0.056 & 0.415 & 4.788 & 0.078 & 0.974  \\ 
        \checkmark & \checkmark &\textbf{0.048} &\textbf{0.315} & \textbf{4.172} & \textbf{0.068} & \textbf{0.982}  \\
        \hline
     \end{tabular}
    } 
   
\end{table}

\begin{table}[t]
  \centering
    \caption{The ablation study on the vector initialization.}
    \label{tab8}
    \resizebox{0.48\textwidth}{!}{
    \begin{tabular}{c |c c c c c } 
    \hline
    $a$, $b$ Init. & Abs Rel$\downarrow$ & Sq Rel$\downarrow$ & RMSE$\downarrow$ & RMSE log$\downarrow$ & $\delta \uparrow$ \\ \hline
        Uniform &  0.050 & 0.333 & 4.328 & 0.070 & \textbf{0.983}  \\
        Kaiming Normal &  0.050 & 0.336 & 4.201 & 0.071 & 0.976  \\ 
        Kaiming Uniform &\textbf{0.048} &\textbf{0.315} & \textbf{4.172} & \textbf{0.068} & 0.982 \\ 
        \hline
     \end{tabular}
    } 
   
\end{table}

\begin{table}[t]
  \centering
    \caption{The ablation study on the number of Res-DSC modules.}
    \label{tab6}
    \resizebox{0.48\textwidth}{!}{
    \begin{tabular}{c |c c c c c } 
    \hline
    Num of Res-DSC & Abs Rel$\downarrow$ & Sq Rel$\downarrow$ & RMSE$\downarrow$ & RMSE log$\downarrow$ & $\delta \uparrow$ \\ \hline
        2 &  0.052 & 0.358 & 4.420 & 0.073 & 0.975  \\
        3 &  0.052 & 0.345 & 4.360 & 0.073 & 0.976  \\ 
        4 &\textbf{0.048} &\textbf{0.315} & \textbf{4.172} & \textbf{0.068} & \textbf{0.982} \\ 
        6 & 0.050 & 0.344 & 4.284 & 0.070  & \textbf{0.982}  \\
        \hline
     \end{tabular}
    } 
   
\end{table}

\begin{table}[!t]
  \centering
    \caption{The results of pose estimation.}
    \label{tab7}
    \resizebox{0.48\textwidth}{!}{
    \begin{tabular}{c | c| c |c} 
    \hline
    Method	& ATE (Seq 1)$\downarrow$ & ATE (Seq 2)$\downarrow$ & ATE (Seq 3)$\downarrow$\\
    \hline 
      EndoDAC \cite{cui2024endodac}  &0.0521 &0.0390 &0.0779\\
      Ours  &\textbf{0.0505} &\textbf{0.0371} &\textbf{0.0771}\\
    \hline
    \end{tabular}
    }
\end{table}

\subsection{Pose Estimation}
In unsupervised monocular depth estimation, camera pose estimation and depth estimation are jointly learned tasks, where the accuracy of the pose network's estimations can also indirectly reflect the precision of the depth estimation. Therefore, we also evaluated the performance of the pose network on the SCARED dataset. We assessed three video sequences using the Absolute Trajectory Error (ATE) metric, comparing our results only with the most competitive method, EndoDAC \cite{cui2024endodac}. The quantitative results are presented in Table \ref{tab7}, where our method outperforms the compared method across the three sequences. The visual results, shown in Fig. \ref{fig5}, demonstrate reduced trajectory drift and a closer alignment with the ground truth for our method. This further validates the effectiveness of our depth estimation approach.

\section{CONCLUSIONS}
In this work, we apply foundation models to the task of endoscopic image depth estimation. To address the challenge of limited medical image data and the inability to train foundation models from scratch, we propose a low-rank adaptation module based on random vectors to fine-tune existing depth estimation foundation model. We introduce a residual block based on depthwise separable convolution to enhance the network's capability to capture local features. Comparative experiments show that our method achieves state-of-the-art performance on the SCARED dataset with the minimum number of trainable parameters. The experimental results on the Hamlyn dataset further validate the strong generalization ability of our method. Ablation studies validate the effectiveness of the proposed modules, and the favorable results from the pose estimation experiments further reflect the accuracy of our approach. Integrating this method into augmented reality navigation systems for minimally invasive endoscopic surgery could enhance surgeons' spatial awareness of the internal anatomy, thereby improving the safety and precision of the surgical procedure.

Meanwhile, our method has certain limitations in inference time. The large number of parameters in the foundational model results in longer inference times compared to conventional methods. Through testing, our method achieves an average inference time of 30ms per image, with a frame rate of 33 frames per second, which is sufficient to meet real-time requirements. Future work could explore further optimization of the network architecture to achieve more efficient depth estimation.






{
\bibliographystyle{ieeetr}
\bibliography{root}
}

\end{document}